%% file: Main.tex
    \def \paperTitle {ABCTracker: an easy-to-use, cloud-based application \\ for tracking multiple objects}

    \documentclass[10pt,twocolumn,a4paper]{article} 
    
    \usepackage{cvpr}
    \usepackage[utf8]{inputenc}
    \usepackage{times}
    \usepackage{epsfig}
    \usepackage{graphicx}
    \usepackage{amsmath}
    \usepackage{amssymb}
    
    \usepackage{natbib}
    \usepackage{graphicx}
    
    \usepackage{enumitem}
    \usepackage{fontawesome}
    
    \usepackage{epstopdf}
    \usepackage{array}
    \usepackage{multirow}
    \usepackage{rotating}
    \usepackage{color}
    \usepackage{colortbl}
    \usepackage{tabulary}
    \usepackage[export]{adjustbox}
    \usepackage[pagebackref=true,breaklinks=true,colorlinks,bookmarks=false]{hyperref}
    \usepackage{etoolbox}
    \usepackage{xargs}                      
    
    \usepackage{nomencl}
    
\definecolor{LightCyan}{rgb}{0.88,1,1}

\cvprfinalcopy %
\ifcvprfinal\fi

\begin{document}

        \title{\paperTitle}
        \author{
            Lance Rice$^1$,
            Samuel Tate$^1$,
            David Farynyk$^1$,
            Joshua Sun$^1$,
            Greg Chism$^2$, \\
            Daniel Charbonneau$^2$,
            Thomas Fasciano$^1$,
            Anna Dornhaus$^2$,
            and Min C. Shin$^1$\\
            \begin{tabular}{c c}
                $^1$~University of North Carolina at Charlotte &
                $^2$~The University of Arizona\\
            \end{tabular}\\
            }

        \maketitle
        
        \begin{abstract}
            Visual multi-object tracking has the potential to accelerate many forms of quantitative analyses, especially in research communities investigating the motion, behavior, or social interactions within groups of animals. Despite its potential for increasing analysis throughput, complications related to accessibility, adaptability, accuracy, or scalable application arise with existing tracking systems. Several iterations of prototyping and testing have led us to a multi-object tracking system – ABCTracker – that is: accessible in both system as well as technical knowledge requirements, easily adaptable to new videos, and capable of producing accurate tracking data through a mixture of automatic and semi-automatic tracking features. \newline \url{http://ABCTracker.org/}
        \end{abstract}

\section{Introduction}
    
    Automated visual tracking systems provide an indispensable tool for gathering reproducible data on groups of animals. Relative to manual analysis and the effort it entails, tracking systems enable researchers to collect quantitative data in ways not previously considered and at more significant scale. To fully recognize the usefulness of tracking systems, it is essential to realize that these systems are almost always a means to an end, rarely the end itself. Dell et al. \cite{dell2014} describes the typical procedure for image-based hypothesis testing as three interdependent steps: imaging, tracking, and analysis. From this perspective, it is apparent that the goal of tracking systems is to enable an effective and efficient transition from video recordings to analysis. Obtaining such a goal in a generalized manner has many challenging facets related to system design and the tracking algorithm.

\input{SystemOverview.tex}

    When individuals seek to gather trajectory data using a tracking system, the first complication they can encounter relates to accessibility. Prominent elements of accessibility include availability, system requirements (e.g., GPU, RAM, CPU), technical knowledge requirements (e.g., experience applying image processing techniques), as well as the overall usability of the application interface. Still, a highly accessible system can fall short of the individuals' needs.  For example, a lack of necessary system features  (e.g., sufficient tools for correcting tracking errors) can limit a system's applicability. Furthermore, scaling issues can become a problem when attempting to process many video sequences or recordings over long periods.

    Although closely related, additional challenges exist regarding the design of the tracking algorithm.  Typical difficulties faced by tracking algorithms include similar appearances between different objects, crowded scenes, and prolonged occlusions among a sometimes unknown and fluctuating number of objects. Systems seeking to generalize to a range of target types must account for difficult to predict motion patterns as well as abrupt pose variations.  Additionally, complex backgrounds, possibly containing clutter, can result in similarities in color, texture, and shape to the objects of interest. As pointed out in \cite{idtrackerai}, lighting also plays a crucial role in the tracker's ability to perform.

\section{System Overview}
    The current version of ABCTracker represents several iterations of prototyping and testing in collaboration with biologists researching social insects. This process has yielded a multiple object tracking system that is highly accessible and can adapt to new videos with virtually no technical knowledge requirements on the user. The system guides individuals through a three-step procedure (Mark-Track-Correct) towards gathering tracking data (figure-\ref{Teaser}). First, the user marks the targets in a small number of video frames (section-\ref{setupsection}). These annotations provide all the information needed to tune tracking parameter values automatically. The second phase, tracking, determines the appropriate parameter values for tracking and applies a sequence of subroutines to construct trajectory data (algorithm outline shown in figure-\ref{Flowchart}). Finally, the system presents the results to the user in the correction phase (section-\ref{correctionsection}). The correction phase includes two modes of correction: a manual mode that offers a complete kit of correction operations, and a guided mode which aims at providing an intuitive and effective means of addressing common tracking errors. 
    
    ABCTracker uses a client/server architecture to reduce users' machine system requirements, facilitate batch processing of videos, and promote flexibility in terms of future development.   Additionally, ABCTracker uses a modular tracking architecture that allows for substituting subroutines individually or replacing the entire tracking pipeline. Using a modular architecture allows the system to adapt as developments continue within the tracking research community.  In this paper, we describe the default tracking pipeline, which was developed as a general-purpose multi-object tracking algorithm for static camera recordings.  So far, the default tracking pipeline has been successfully applied to more than 350 videos of both marked (e.g., painted color markers) as well as unmarked objects (examples provided in figure-\ref{Gallery}). Section-\ref{EvaluationSection} provides tracking performance examples on various types of recordings. In addition, we demonstrate the effectiveness of our two modes of correction -- guided mode, and manual mode -- to visualize and address errors that occurred during tracking.

\section{Related Works}
    The user interaction workflows within related works follow a similar multi-step procedure. The first step relates to setting up the tracking algorithm (also called initializing or tuning), followed by applying the tracker. In some cases, the system also includes a correction step (sometimes referred to as validation). ABCTracker also follows this multi-step procedure which we describe as the Marking, Tracking, and Correction phases.

\input{Algorithm_setup_comparison.tex}

    When discussing various approaches taken to the setup step, it is important to note that existing tracking systems rely heavily on foreground images (also referred to as foreground blobs) \cite{idtracker,idtrackerai,umatracker,bemovi,toxtrac,ctrax,biotracker}. Because of this, several of the parameters a user tunes during this step relate to that fact.  In many cases, when a tracking system completely fails or produces low accuracy results, it is due to the quality of foreground estimation or violated assumptions on its properties. For example, IdTracker \cite{idtracker} and IdTracker.ai \cite{idtrackerai} both employ fingerprinting techniques to distinguish targets from one another; this imposes particular assumptions on the foreground needed for gathering training examples. Additionally, nearly every system available performs inference on the foreground for gathering object detections. Thus, the tools provided for tuning foreground estimation are crucial and, roughly speaking, fall into two categories. The first category corresponds to predefined image processing pipelines (such as \cite{idtracker}). These approaches present several parameters of foreground estimation as well as one or more sample images so the user can try various parameter settings.  The second category of systems provides further flexibility by allowing the user to build custom image processing pipelines (such as \cite{SwisTrack,umatracker}). The primary disadvantage in both categories is technical knowledge requirements. Such requirements include experience using image processing techniques (e.g., element size for binary erosion) as well as considerations on how trade-offs in qualitative properties of the foreground relate to the algorithm's ability to track (e.g., over-segmentation vs. under-segmentation). ABCTracker proposes a different approach that not only tunes the parameters of foreground estimation but all necessary parameters used for tracking and error correction (section \ref{setupsection}).

    Several works use sequential tracking methods (i.e., may also be referred to as online or real-time) which employs an inference procedure that sequentially processes video frames one at a time \cite{umatracker, toxtrac, biotracker}. A notable advantage to these approaches is their innate appropriateness for real-time application.  The drawback of these approaches is that their observation window is limited. For example, occlusions can be challenging to manage because no information (i.e., observations/detections) is available relating to the other end of the occlusion. Batch approaches (also referred to as offline, or Global data-association based tracking approaches) express tracking as an optimization problem over larger temporal windows of observations than typically considered in sequential approaches. Two systems that have successfully employed batch tracking approaches are IdTracker \cite{idtracker} and IdTracker.ai \cite{idtrackerai}.  Here, they extract discriminative features from the targets which serve as fingerprints for performing data association. Both works require a sufficient number of training examples to work, meaning that several assumptions are placed on the recording conditions, cleanliness of the background, and the density of targets in the scene. ABCTracker formulates tracking in a batch manner but uses predictions from multiple sequential trackers to conservatively perform data association (see section-\ref{trackingsection}).  

    Some systems implement correction features to handle errors produced by the tracker \cite{umatracker, biotracker, idtrackerai, toxtrac, bemovi, ctrax}. UMATracker \cite{umatracker} incorporates correction abilities during tracking by allowing the user to monitor and pause the tracking algorithm.  While paused, the user can either adjust target positions or swap id assignments and then resume tracking. IdTracker \cite{idtrackerai} proposes a crossing-validation step that asks the user to verify the presented id assignment during an animal crossing (i.e., an occlusion) -- no other correction operations are available (e.g., adjustments). A few systems exist that offer more complete correction operation sets \cite{biotracker, ctrax}, but lack intuitive user interface designs. ABCTracker includes two integrated modes of performing corrections: a comprehensive correction mode we refer to as manual mode, and an assisted error correction mode called guided mode (see section-\ref{correctionsection}).

\section{Marking phase}
    \label{setupsection}
    Multi-object tracking algorithms are formulated as pipelines of subroutines (e.g., preprocessing images, detection, low-level association, filtering). Each subroutine has a set of parameters that impact not only its performance but the performance of any subroutines that follow. This makes the initial phase within tracking systems - parameter tuning (also referred to as setup or preprocessing) - crucial for the overall success of the system. Unfortunately, in previous tracking systems, parameter tuning is also very demanding on users in terms of technical know-how, sometimes so much as to have the user assemble both the tracking pipeline and its parameter values directly (e.g., ForegroundThreshold = 0.9). An ideal system should allow someone with no technical knowledge of how the tracking algorithm works to tune all subroutine parameters effectively.
    
    Our tracking system requires only the input video and user annotations on select video frames (which we will refer to as user-marked-frames); no tracking parameters need to be defined directly by the user. A user-mark (figure \ref{Teaser} and figure-\ref{UserMarks}) is a region of the image corresponding to a target of interest. Three-clicks, together with brush size, define the extent of a target. A smooth spline is fit to the three points, and the resulting thickness of the spline is equal to the brush size. The system requests user-marks (i.e., three-point splines with a defined thickness) for a number of video frames determined as follows. The first and last frames of the video are always requested. Knowing the location of each target in the initial and final video frames is beneficial to both the method used during tracklet matching (section-\ref{matchingsection}) as well as guided correction mode (section-\ref{guidedModeSection}).  If the video is longer than 5000 frames, then chunking will be performed (see section-\ref{chunkingsection}), and the system requests user-marks for the overlapping frame between neighboring chunks. If the previous two conditions are met, but less than 30 user-marks have been gathered, more frames are requested for marking. The additional frames are selected uniformly throughout the video until 30 marks have been gathered.

\input{AlgorithmFlowChart.tex}

\section{Tracking phase}
    
    \label{trackingsection}

    \subsection{Foreground Estimation}
        \label{foregroundsection}

        Currently, ABCTracker assumes input videos are captured using a static camera setup (i.e., no camera motion). The foreground pixels are initially estimated with background subtraction using the user-marked-frames to determine the proper difference threshold. Because the initial foreground estimation can contain varying levels of noise and produce overly connected and fragmented foreground blobs, we use particle swarm optimization on a set of morphological operations for foreground refinement. We use four parameters for foreground refinement. The first parameter defines an initial threshold for the minimum area foreground blobs should have. The second parameter specifies the number of times to perform the majority operation. The final two parameters concern the size of the structured element for performing morphological closing, and again the minimum area threshold on the foreground blobs resulting from the previous operations. We formulate the loss function of the particle swarm as a weighted average of the number of correct, over-segmented, under-segmented, and incorrect foreground pixels. All factors within the particle swarm loss are calculated using user-marked-frames. 

    \subsection{Detection}
        \label{detectionsection}
        The detection module produces location, size, and orientation information of detection responses in each frame. First, the system trains an SVM classifier that operates on HOG features \cite{dalal2005histograms} extracted from blob proposals. User-marks determine positive and negative examples for training the classifier. Using the foreground blobs as region proposals reduces the frequency of false-positive detections. After foreground estimation and refinement are applied to a given frame (section-\ref{trackingsection}), foreground blobs that meet the minimum and maximum thresholds for both area and ratio (determined using user-marked-frames) are given to the previously trained classifier. Proposals receiving positive classifications become detections. The set of detections across all frames of the video, represented as $D$, proceed to tracklet building. Note that, when the average size of the targets within the scene is small (detection area $<50$ pixels), HOG features become unsuitable for classifying blob proposals. Therefore, when the average target size is less than 50 square pixels, the classifier is not used, and blob proposals that pass all area/ratio thresholds become detection responses.

    \subsection{Tracklet building}
        \label{trackletbuildingsection}

        To assemble low-level tracklets (i.e., confident trajectory segments constructed from frame detections), we develop an approach that can accurately generate long detection associations without relying on motion/appearance features or learning function parameters for scoring associations. We accomplish this by performing inference on the foreground images, specifically foreground "tunnels" that span spatially and temporally over the video sequence.  We construct the foreground tunnels according to Fasciano et al. \cite{fascianoTunnels}, which results in a directed acyclic graph $F = (V, E)$ that represents overlapping foreground blobs through time. 
        
        Expected inputs to tracklet building are the set of detections $D$ and the foreground tunnels $F = (V, E)$. We first determine regions of the foreground tunnels $F$ that neither merge nor split.   For every $v_i \in V$ (i.e., foreground blob) we calculate its in-degree, $deg_-(v_i)$ (number of inbound edges), and out-degree, $deg_+(v_i)$ (number of outbound edges). Sequences of nodes having $deg_-(v_i)=1$ and $deg_+(v_i) = 1$ partition the graph $F$ into “lanes” which represent overlapping foreground blobs across time that neither merge nor split. We then map detections in $D$ to foreground blobs in $v_i \in V$ by determining if the centroid of the detection $d_u \in D$ falls within the boundaries of blob $v_i \in V$. We convert lanes into tracklets when a majority of their blobs have a detection mapped.   Blobs within a lane having no mapped detection but are part of a majority mapped lane are assigned interpolated values. We discard all detections present in non-majority lanes. The output of tracklet building is the initial set of tracklets $T_0$.

        The final subroutine before tracklet matching (section-\ref{matchingsection}) concerns learning a more robust measure of detection confidence. The SVM classifier described in section-\ref{detectionsection} is very limited in terms of training data; thus, its prediction confidence is unreliable. We learn a Random Forest classifier using training examples from the initial tracklet set $T_0$. Negative examples are gathered by taking detections in $T_0$ and adding random noise in position and orientation. This confidence measure is used during tracklet matching as well as during the correction step, but initially, it is used to filter false positive tracks within $T_0$. We remove every track $t_i \in T_0$ that has average confidence less than 0.5.

\input{Forward-backward_GAPF.tex}

    \subsection{Tracklet Matching}
        \label{matchingsection}

        The goal of tracklet matching is to establish associations between tracklets within the initial set $T_0$. Instead of attempting to estimate tracklet affinities with parametric heuristics or association models learned offline with training data or online with restrictive assumptions imposed, we rely on prediction agreements between independent sequential trackers. The underlying assumption is that accurate tracking should be independent of the temporal direction and that consistencies between predictions gathered in opposing directions signify trustworthy associations. Figure-\ref{IterativeMatching} illustrates the general concept behind the matching algorithm. We first describe how the sequential tracker operates and what it means to run the tracker in both the forward and backward directions. Afterward, we outline how we use prediction agreements between the forward/backward trackers to perform matching.  
        
        \paragraph{Forward/backward sequential trackers} 
            We describe how we apply the sequential tracker in the forward direction; except for temporal direction, the approach is identical in the backward direction. Inputs are the current set of tracklets $T_n = \{\tau_i\}$ and the set of foreground pixels in each frame $\{F_t\}$.  The sequential tracker seeks to find a global configuration for all targets in the scene that maximizes both appearance similarity (i.e., image patch comparisons) and foreground coverage. For a particular frame $f_t$, estimating the global configuration of the targets in that frame means: estimating the location of tracklets which terminated before frame $f_t$, and factoring in tracklets having known locations within frame $f_t$. To factor in the tracklets having known locations in frame $f_t$, we remove foreground pixels from $F_t$ corresponding to those locations. Specifically, for every tracklet within frame $f_t$, we determine pixels within the tracklet's oriented bounding box location and subtract these pixels from the foreground image $F_t$, producing ${F}^{\star}_t$. For each frame of the video $f_t$, the general procedure the tracker follows is: 1.) it determines tracklets which terminate in frame $f_{t-1}$ and adds them to the list of targets to track $\Psi$, 2.) it calculates ${F}^{\star}_t$ by removing foreground pixels from $F_t$ corresponding to known target locations, 3.) it optimizes the configuration of targets being tracked $\Psi$, 4.) it records if any tracks within $\Psi$ "landed" on a tracklet which begins in frame $f_t$ (i.e., records targets in $\Psi$ that are sufficiently close to the starting frame position of a tracklet $\tau_i \in T_n$ -- represents a prediction), and 5.) it removes any targets that "landed" from the list of targets to track $\Psi$.  We determine the distance threshold for checking if a track landed based on the user-marks (section-\ref{setupsection}). 
            
            Because tracks can fail to land (e.g., drifted to background), the set of targets to track $\Psi$ could grow very large and result in unnecessary computation. Using the user-marked frames, we determine an upper bound on the number of targets in the scene $\Omega$. After processing each frame during sequential tracking, we remove $q$ targets from $\Psi$. Specifically:
            \begin{equation}
                q = |inFrm(T_n, f_t)| + |\Psi| - (\Omega + 2)
            \end{equation}
            where $inFrm(T_n, f_t)$ returns the set of tracklets occurring in frame $f_t$. When $q>0$, we determine q members within $\Psi$ to terminate by calculating a cumulative version of the target-level fitness for each target (equation-\ref{targetFitnessEQ}). In this way, targets who remain in $\Psi$ longer are more likely to be part of the $q$ members terminated. 
        
        \paragraph{Optimization}
            We now describe how the global configuration of targets to track is optimized in each frame using a genetic algorithm approach.  Here, each member in the genetic algorithm's population, $\hat{\Psi}_j \in \mathbf{\hat{\Psi}^n}$, represents a potential configuration of the targets to track, $\hat{\Psi}_j = \{\hat{\psi}_i\}$. The initial population $\mathbf{\hat{\Psi}^0}$ is constructed by adding random displacement ($\mathcal{X} \sim \mathcal{N}(0, \sigma^2)$) to target locations in the previous frame. Motion statistics used during population initialization and reproduction are calculated from the initial set of tracklets, $T_0$. To estimate the optimal configuration of target positions, we use two fitness terms: a target-level fitness $Fit_{T}(\hat{\psi}_i)$ and a global configuration-level fitness $Fit_{G}(\hat{\Psi}_j)$. The target-level fitness calculates the appearance similarity at a predicted location for a single target $\hat{\psi}_i \in \hat{\Psi}_j$:
            \begin{equation}
                \label{targetFitnessEQ}
                Fit_{T}(\hat{\psi}_i) = \Bar{F}_{\Delta}(\Delta(\hat{\psi}_i))
            \end{equation}
            
            Where $\Bar{F}_{\Delta}$ is a complementary cumulative distribution function and $\Delta(\hat{\psi}_i)$ is the absolute difference between the oriented color image patch at position estimate $\hat{\psi}_i$ and the target's appearance template. We determine target appearance templates based on detection confidence (see section-\ref{trackletbuildingsection}). When using multiple templates, we select the template with the minimum absolute difference.  The system calculates the sample mean and standard deviation of $\Bar{F}_{\Delta}$ at the start of each tracklet matching iteration. Specifically, for every tracklet $\tau_i \in T_n$ we compute $\Delta$ at randomly selected locations within $\tau_i$. Global configuration-level fitness concerns how well a predicted configuration explains the unclaimed foreground within ${F}^{\star}_t$:
            \begin{equation}
                \label{globalFitnessEQ}
                    Fit_{G}(\hat{\Psi}_j) = Cov(\hat{\Psi}_j, {F}^{\star}_t)
            \end{equation}
            
            where $Cov(\hat{\Psi}_j, {F}^{\star}_t)$ is the number of foreground pixels in ${F}^{\star}_t$ covered by predicted locations $\hat{\psi}_i \in \hat{\Psi}_j$. Fitness scoring and reproduction is performed over $\omega$ cycles. The output configuration is the weighted mean of all members within the final population, $\mathbf{\hat{\Psi}_\omega}$. Specifically, for each target to track $\psi_i$, the final state estimate is:
            
            \begin{equation}
                \sum_{\Psi_j \in \mathbf{\Psi^{\omega}}} {(1-\mathcal{W}_{ij}) \hat{\psi}_{ij}}
                \label{PopulationMeanEquation}
            \end{equation}
            
            where $\mathcal{W}_{ij}$ represents the normalized target-level fitness values (equation-\ref{targetFitnessEQ}).

        \paragraph{Reproduction}
            We first select two parents by roulette wheel selection based on global configuration-level fitness scores (equation-\ref{globalFitnessEQ}). Given two parents, a single child is produced by randomly performing either crossover or mutation. In the case of crossover, we iterate over each target being tracked and assign the highest scoring predicted location (equation-\ref{targetFitnessEQ}) among the two parents to the child. For mutation, we first select the parent with the highest configuration-level fitness (equation-\ref{globalFitnessEQ}) and add random displacement on position and orientation to all targets being tracked. As previously noted, displacement parameters are calculated from the initial set of tracklets. 
        
        \paragraph{Iterative Tracklet Matching} 
            Each iteration of matching $n$ uses the previous tracklet set $T_{n-1}$ to produce a further associated tracklet set $T_n$. During each iteration, we construct a directed association graph $G = (V, E)$. Each tracklet is represented by two vertices corresponding to the beginning and end of the tracklet -- no edge is defined between the two endpoints. Directed edges in $G$ capture predictions from the sequential trackers (i.e., targets to track which "landed"). We first detect prediction agreements by finding graph cycles with a length of 2. We then filter all detected cycles which have a vertex with more than one inbound edge. A vertex with more than one inbound edge would mean that multiple predictions "landed" on that tracklets' endpoint, and thus would represent an unreliable association. Finally, we associate the tracklet endpoints corresponding to the filtered cycles detected in $G$. After each matching iteration, appearance models of connected tracklets are updated to reflect the most representative appearance template. In practice, matching continues for three iterations or until no connections are made in a single iteration.

    \subsection{Automatic video chunking}
        \label{chunkingsection}
        As a step towards more scalable application, the system employs automatic video chunking and stitching for videos of sufficient length. Aside from additional frames requested during the marking phase, the process of video chunking and stitching is hidden from the user. For videos having more than 7000 frames, the system determines frame boundaries for each chunk according to two system parameters: the ideal number of frames per chunk  $\lambda$ and the minimum number of frames per chunk $\lambda_{min}$. Furthermore, we define the boundaries for each chunk such that temporally neighboring chunks have precisely one overlapping frame in common. Currently, we set the values for $\lambda$ and $\lambda_{min}$ to be 5000 and 300 respectfully. These values represent the typical sequence length used during algorithm development and also result in an acceptable amount of memory use. We add the overlapping frames between neighboring chunks to the list of user-mark frames requested during the marking phase. Doing so provides the system with location information for all targets within these frames. 
        
        Once the tracking phase has completed all of the defined tracking modules, the final routine is to aggregate all chunk results into a single tracklet state.  The system performs nearest neighbor matching to stitches the tracklets across all chunks. Association errors resulting from nearest neighbor stitching are practically impossible since target locations are known within the overlapping frames.

\section{Correction phase}
    \label{correctionsection}

\input{Large_GuidedInferenceExample.tex}

        Here we describe the correction phase of a tracking process where the user can locate and address tracking errors. ABCTracker's correction system includes two modes: manual mode and guided mode. Manual mode offers multiple ways of visualizing all tracks within the video and provides a complete set of track modification operations for obtaining high accuracy results. Guided mode provides a directed work-flow by automatically determining possible errors in the results and presenting them to the user as easy to answer questions (which we will refer to as \textit{reviews}). The two correction modes have been designed to be quick and easy to switch between at any time, promoting flexible work-flows suitable for novice as well as advanced users.

        \subsection{Manual correction mode}
        
            One of the primary design goals for manual correction mode is providing a fundamental set of correction operations. As illustrated in figure-\ref{Gallery}, there exist a large number of ways in which target objects can behave in addition to numerous approaches one can take towards assembling, confining, and recording them. As such, the correction phase needs to be able to handle cases where the tracker produces poor results while also offering features to accelerate the correction process when the tracker performs reasonably. We include five correction operations within manual mode to handle any tracking error (Table-\ref{Error-type-list}): Add, Remove, Join, Break, and Adjust. 
    
            One noteworthy implementation detail is that each operation within manual mode has a predictable behavior. For example, the break operation slices the selected track precisely at the current frame and modified positions during an adjustment operation use linear interpolation. Previous prototypes of the manual correction mode attempted to perform inference on operation requests (e.g., break operation would attempt to find nearby associations with high uncertainty).  When using these versions of the system, users expressed frustration towards not having manual operations that were predictable in their outcome. 

        	The system includes several features reported as being useful by users. For example, target-following zoom controls, playback speed controls, autosave, an undo operation available in both correction modes, keyboard shortcuts, video scrubbing with the mouse wheel, and context-aware help dialogs. Moreover, two visualization tools are available to aid in finding tracking errors favored by the algorithm.  When the user selects a track, an interactive visualization appears above the playback bar (item d within figure-\ref{ManualVisualizationFeatures}) depicting the temporal extent of the selected track. Similarly, the temporal extent of all tracks within the video can be interacted with (item c within figure-\ref{ManualVisualizationFeatures}), allowing the user to jump to potential false positive tracks and identify false negative associations quickly.

\input{Tracking_Gallery.tex}

\input{PossibleErrorsTable.tex}

\input{Corrections_Interface.tex}

        \subsection{Guided correction mode}
            \label{guidedModeSection}
           Guided-mode aims at providing an intuitive and effective means of addressing two common error types within section-\ref{EvaluationSection}: FN associations, and ID-integrity errors. The two error types are related – one being a failure to associate tracklets and the other an incorrectly made association – and represent a trade-off made while handling uncertainty. While the tracking algorithm (section-\ref{matchingsection}) attempts to establish as many correct associations as possible, the key principle in its design is to avoid ID integrity errors and instead favor FN associations. This decision is due to the innate dependency between tracking and error correction. More specifically, tracking should approach the trade-off with the following realities considered. First, FN associations are easier to detect (manually as well as automatically) and more intuitive for users to correct than ID-integrity errors. Second, correcting situations that involve an ID integrity error are often multi-step processes (i.e., just identifying an incorrect association will result in an FN association error unless the user provides additional information). Lastly, the act of correcting FN associations has the potential to identify ID integrity errors (see figure-\ref{GuidedInference_Large}).

            \subsubsection{Work-flow of guided reviews}
            A guided-mode review represents a possible tracking error identified in the results. The process a user follows for answering a review is identical for all reviews, regardless of the underlying error the user is addressing (illustrated in figure-\ref{GuidedInference_Large}). First, the user is presented with a target to follow and prompted to begin video playback. From there, playback will periodically pause at predefined frames, which we will refer to as \textit{keyframes}, and ask the user to annotate the target's location and orientation with a click-n-drag interaction. This process continues until the review has finished, at which point the review's answer is sent to the server for inference while the next review is presented (details in section-\ref{correctionImplementationSection}).

            \subsubsection{Review creation and ordering}
                \label{ReviewCreation}
            
                Three properties of a review dictate the requested interaction with the user: the ID of the tracklet being reviewed, the starting frame of the review (i.e., step showing what target to follow), and the key-frames selected for annotation. Because many review situations relate to occlusions or large pose variations, it may be difficult to discern what target to follow. Because of this, we select the starting frame of the review based on detection confidence (described in section-\ref{trackletbuildingsection}) and allow the starting frame to be within 45 frames of the believed starting point of the error. 
                
                As stated previously, all reviews within guided mode appear to be similar from the user's perspective and are applied using a common inference procedure (section-\ref{ReviewInference}). We make a distinction here between review types to better describe the creation and ordering of reviews. 
                
                Fragment reviews primarily address FN associations. FN associations can be detected with perfect recall by searching for tracks that end before the last frame of the input video. The precision of detecting these failures using this approach is affected by the frequency of false-positive tracks and targets exiting the scene. We note that these two cases can be handled during guided reviews with the options to remove the track or state that the target is no longer in the scene. The system prioritizes fragment reviews in ascending order based on the starting frame of the review. 
                
                For connection reviews, which determine possible ID assignment issues incurred during tracking, a more detailed process is required. Individually reviewing all connections made throughout the tracking process is impractical for nearly any tracking situation. Furthermore, the number of ID assignment errors made should be relatively small compared to the number of associations established; for the tracking algorithm described in section-\ref{matchingsection}, we attempt to justify this assumption even more so through design decisions which favor FN associations over ID integrity errors. Therefore, we utilize a moving threshold on priority scores for connection reviews. A moving threshold allows the system to request more connection reviews as ID assignment issues are identified. The priority value for connection review is the average detection confidence (section-\ref{trackletbuildingsection}) within the connected region (i.e., between the endpoints of the association). 
            
            \subsubsection{Inference on answered reviews}
                \label{ReviewInference}
                
                The general concept behind review inference is that key-frame annotations for select tracklets can be used to join disconnected tracking segments. Additionally, the same procedure can check for incorrect associations by first breaking any connections the user reviews and then relying on the key-frame annotations to reestablish correct associations.
                
                Let $r$ and $\tau_i$ be the review and the tracklet being reviewed by the user. Let $r_{start}$ and $\mathbf{K} = \{k_j\}$ be the starting frame of the review and the set of key-frame annotations respectfully. We record every association made on a tracklet during tracklet matching (section-\ref{matchingsection}) within the properties of the tracklet. Using this connections property, we determine if the temporal span of the annotation set, $\mathbf{K}$, covers any connection. If so, then the review concerns an uncertain association made during matching, and we break the connection. After breaking connections within $\tau_i$ covered by $\mathbf{K}$, we apply the annotations to the tracklet. The position and orientation information for each key-frame annotation is assigned to the tracklet, and the following interpolation procedure fills the gaps between the key-frame annotations. For each consecutive pair of key-frame annotations, we run a single-target particle filter both forward and backward. If both tracking directions "land" on the opposite key-frame annotation, then agreement was established (similar to section-\ref{matchingsection}). In this case, we assign the weighted mean of the two trajectories as the interpolated positions. We use linear weights, from 1 to 0, when combining the sequential tracker predictions such that more weight is given the starting point of the trackers. If either of the trackers fails to "land," then we increase the number of particle samples used and rerun tracking. If the tracking directions do not both "land" after four attempts, then we resort to a fallback and assign linear interpolation as the interpolated values for the pair of key-frames. 
                
                In some cases, the key-frame annotations within a single review will not provide enough information to join two tracklets confidently, thus requiring the review to be re-queued for additional annotations (see figure-\ref{GuidedInference_Large}). So, before determining if the reviewed tracklet should be joined with another tracklet, we gather any previous key-frame annotations that this review builds upon; let $\mathbf{K^\star}$ be the set of previous and current key-frame annotations. 
                
                We use a sequence of four checks to determine if a reviewed tracklet should be joined with another. First, we calculate the distance between key-frame annotations and tracklets other than $\tau_i$. The first threshold checks that the average distance between tracklets and annotations are less the $\theta_1$. Let $\tau_k$ be the closest distance tracklet which passes threshold $\theta_1$. The second and third thresholds ensure that the join operation is conservative, namely that the margin between the second-best distance is greater than $\theta_2$ and that the number of key-frames temporally overlapping $\tau_k$ is greater than $\theta_3$. A final check eliminates any join operations that have key-frame annotation extending beyond the tracklet to join with, $\tau_k$. The values for $\theta_1$ and $\theta_2$ are determined automatically based on the user-marks. 
                
                The last step of review inference is to identify tracklets that are \textit{engulfed} by the key-frame annotations. We define an engulfed tracklet as a track that is covered both spatially and temporally by the annotations, $\mathbf{K^\star}$. We first calculate the average distance for each tracklet falling entirely within the temporal span of $\mathbf{K^\star}$. Since engulfed tracks represent redundant data and redundant data is harmful to guided corrections (i.e., can cause leap-frogging review situations), we use a relaxed maximum distance threshold of $2 * \theta_1$. We assume annotations are a more valuable option to keep and remove engulfed tracklets.

            \subsection{Correction phase implementation details}
                \label{correctionImplementationSection} 
                In terms of implementation, exchanges between the client and server can be thought of as review-batches and apply-batches. At the start of the correction phase, the server supplies an initial review-batch to the client, specifically an ordered list of guided reviews. Apply-batches from the client to the server contain one or more correction operations (both manual operations and review answers are considered correction operations, meaning a mixture of operation types is possible within a single apply-batch). Each review-batch from the server provides many reviews, allowing the client to continue presenting reviews while the server processes the previous batch. Once the server finishes processing a batch, another round of exchange occurs -- the server creates a new list of reviews, and the client passes it's current backlog of operations to apply. 
                
                Each review-batch constructs an entirely new list of reviews -- a quick process that results in little overhead and aids in maintaining a consistent state of the tracks between the client and server. Since identifying FN association errors can also identify ID assignment errors at inference time, the priority values for all fragment reviews are automatically shifted to be above the highest priority connection review.

\section{Results}
     \label{EvaluationSection}

    \input{GuidedSimulations_Baseline281.tex}

\input{GuidedSimulations_CM1_0693.tex}

    Owing to our collaboration with biologists, ABCTracker has been well tested both in terms of system stability and robustness. The current version of the system has been used to track more than 350 video sequences, constituting over 1.5 million video frames tracked.
    
    We have used ABCTracker to collect ground truth trajectories (position and orientation) for 40 diverse video sequences containing groups of insects. The average number of targets within these videos was 38 and each contained roughly 5000 frames on average captured at 30-fps -- amounting to 6.8 million ground truth positions gathered. Following the marking and tracking phases, ground truth annotators were permitted to use any combination of manual- and guided-mode. To ensure quality in the ground truth and that inference on guided corrections behaved as expected, annotators were asked to perform a final check within manual-mode. The final check consisted of following each target individually throughout the video, correcting any errors detected, and recording the track as complete using the system's “Mark Complete” feature. Manual correction mode proved to be a useful tool for gathering ground truth tracking data. From our informal observations, new users to the system -- not just the annotators previously mentioned -- tend to make higher proportions of their corrections within guided mode than more experienced users do. As an individual becomes familiar with correcting tracking errors and specific features in manual mode (e.g., figure \ref{ManualVisualizationFeatures} – c and d), we see workflows develop that involve more switching between the two modes. Although ABCTracker does not assume that objects are confined to the scene, one limitation within the current version of manual correction mode is that objects which exit and later return to the scene must have different ID values assigned. We want to address this in future versions. 

    \subsection{Evaluation metrics}
        We follow the evaluation procedure outlined in \cite{li2009learning, milan2016mot16} to measure the tracking performance of the system. The first five of the following metrics evaluate the system's ability to detect objects in the scene and filter false-positive responses (i.e., frame-based metrics), while the remaining three metrics assess how well the system associates detection responses:
            
            \begin{description}
            
                \item[Ground Truth Coverage (GTCov. $\uparrow)$] Tracking recall - the percentage of GT positions covered by a tracklet. 
            
                \item[Mostly Tracked (MT $\uparrow$)] The percentage of GT trajectories successfully tracked for more than 80\%. 
                
                \item[Partially Tracked (PT)] The percentage of GT trajectories successfully tracked between 20\% and 80\%.
                
                \item[Mostly Lost (ML $\downarrow$)] The percentage of GT trajectories which are tracked for less than 20\%. 
                
                \item[False alarms per Frame (FAF $\downarrow$)] The number of false detection responses divided by the number of frames
            
                \item[Identity Switches (IDS $\downarrow$)] The number of times a tracklet changes its matched GT identity \cite{li2009learning}.
                
                \item[ID Integrity Errors (IdInteg. $\downarrow$)] The number of times a tracklet changes from one matched GT identity, ${id}_j$, to another, ${id}_k$, and does not return to the former GT identity ${id}_j$ within N frames. Here, we define N to be 30 frames, which is roughly the average frames-per-second, thus one second, in our datasets. 
                
                \item[False Neg. Associations (FN Assoc. $\downarrow$)] The number of times a correct tracklet association was not established \\(i.e., No. of false negative tracklet associations). 
                
            \end{description}
            
        The ID integrity error metric (IdInteg) aims at capturing significant ID assignment issues and can be thought of as a relaxed variant of the ID switch metric described in \cite{li2009learning}. 

    \subsection{Evaluation of tracking and guided correction mode}
        \label{EvalCorrection}
        We evaluate the three major inference performing components within the system: tracklet building (section-\ref{trackletbuildingsection}), tracklet matching (section-\ref{matchingsection}), and guided correction mode (section-\ref{guidedModeSection}). We use two 5000 frame video recordings containing multiple ants for evaluation, each of which have a few distinct challenges (figures-\ref{GuidedModeEval_Baseline281} and \ref{GuidedModeEval_CM1_0693} show sample frames). 
        
        The results presented in the bottom tables of  figure-\ref{GuidedModeEval_Baseline281} and \ref{GuidedModeEval_CM1_0693} show that tracklet building is successfully able to construct a conservative set of initial tracklets. In both cases, detections covering more than 50\% of the available ground truth were assembled into the initial tracklet sets with no ID assignment issues.  Additionally, the average position error (reported in pixels) and false alarm frequency after tracklet building is low and tends to increase as the system performs further stages of inference. This increased error in estimated position is because detection and tracklet building are designed to filter responses corresponding to occlusions and, in general, localization is easier outside of occlusions.  
        
        Figures-\ref{GuidedModeEval_Baseline281} and \ref{GuidedModeEval_CM1_0693} also show that tracklet matching can conservatively establish associations among the initial tracklet set. More than 90\% of the possible associations were successfully established (96\% for \textit{"OpenDivide\_1"} and 91\% for \textit{"PinkArena\_1"}). Between the two videos, a single ID-integrity issue resulted during tracklet matching.
        
        We demonstrate guided correction mode’s ability to detect and correct tracking errors through simulated user input on real video sequences that have ground truth. The simulation was designed to mimic what a user can see and do during a guided correction review. Some noteworthy implementation decisions include: 
        \begin{enumerate}
            \item The annotations are supplied just as they would be for a real user – click-n-drag coordinates – which are determined by finding the closest ground truth target to the track presented. 
            \item The simulation can only use portions of the ground truth that the review (i.e. video playback) has shown 
            \item The simulation is not aware of any other tracks than the one presented
            \item The simulation can use other options that are available within a review just as a typical user can (e.g. target has exited the scene)
        \end{enumerate}
        We reevaluate the tracking performance after each review answer is applied. The results of simulated guided corrections are presented in figures-\ref{GuidedModeEval_Baseline281} and \ref{GuidedModeEval_CM1_0693}. All FN associations, as well as the single Id-Integrity error, were identified and corrected. The two targets reported as mostly lost after answering all guided reviews in \textit{"PinkArena\_1"} were targets that only appeared in the scene for a brief portion of the video and had no tracking. Guided-mode simulations were not capable of adding new tracks, thus was not able to address those errors.

\input{ManualAnnotTable_Baseline281.tex}

        We compare guided corrections with the traditional approach for gathering trajectory data where annotations are defined at set intervals for every object. Specifically, an individual selects a target to follow, watches the video, and annotates the targets position every set number of frames.  The process is repeated for every target in the video. This annotation procedure has been used in video annotation tools such as VATIC \cite{vatic}. Figure-\ref{manualannotationcomparison} presents the comparison results. The traditional approach of manual annotation  achieves similar ground truth coverage as answering all guided corrections when the key-frame step size equals 30 but requires 434\% (1.78 hours) more playback time and 546\% (6,817) more annotations. Figure-\ref{manualannotationcomparison} also reports an overall estimated time which uses the following assumptions and parameters. First, we assume that playback speed has not been altered and that annotations in the manual approach are supplied in the same way reviews are in guided correction mode (i.e., click-n-drag). We also assume that the average time to make a click-n-drag annotation is 1.5 seconds, and starting a new object using the traditional approach as well as starting the next review in the proposed approach takes 2 seconds. Please note that the time parameters for annotations and changing reviews/targets are approximations that may underestimate the total amount of time. Still, since both methods use identical parameters for estimating time, we believe they are representative of the relative difference between the two approaches.

\section{Conclusion and Future Work}
    In this work, we proposed a multi-object tracking system that is very accessible, adaptable, and capable of obtaining accurate tracking data. For adapting the tracking algorithm to new videos, we have proposed an alternative approach that dramatically reduces the technical demand on users compared to prior works. We have included several examples that demonstrate how the initial marking phase successfully provides an intuitive means for tuning all tracking and correction related parameters. Furthermore, the tracking examples provided show that the system can be applied to a variety of video recordings. Specifically, we show the system can track a range of object types under diverse recording conditions with no restrictions as to whether or not the objects can enter and exit the scene. We have also described that the system implements a client/server architecture and automatically chunks long video sequences -- features that attempt to bring the system closer to more scalable application. 
    
    Using two challenging datasets, we have verified the inference performing components within the system. Concerning the tracking algorithm, we have shown that the system can conservatively establish both an initial set of tracklets as well as associations between tracks. We show that guided correction mode can effectively identify potential errors in the tracking results and direct the user through an intuitive procedure of correcting them. Finally, we demonstrate that guided corrections is more efficient than traditional approaches of manually gathering trajectory data. 
    
    In summary, the current version of the system has the following advantages:
    
    \begin{itemize}[label=\faChevronCircleUp]
        \item Easy to get started. Once downloaded, the user can begin right away without any complicated installations or reading of technical manuals. The only dependency is java, all other dependencies are pre-packaged with the client. Furthermore, the application's interface can run on most machines since the system requirements are very low.
        
        \item An overall user-friendly interface with context aware help dialogs that introduce system features as the user encounters them for the first time. 
        
        \item Easy to tune the algorithm. Adapting the tracker to new videos only requires user marks, no direct tuning of image processing pipelines or other tracking related parameters. 
        
        \item The tracker generalizes well for a variety of static camera recordings.  Thus far, it has been applied to a number of target types, both marked and unmarked, within simple as well as more complex habitats. Additionally, the tracker makes no assumptions as to whether objects are confined to the recording area or not. 
        
        \item Automatic chunking allows for tracking of longer video sequences than would otherwise be possible within the system. 
        
        \item The manual correction mode contains operations which can address practically any tracking error and also includes visualization tools to help identify errors.
        
        \item The correction phase includes a guided mode that can effectively identify common tracking mistakes and directs the user through an intuitive procedure of correcting them. 
    
    \end{itemize}

We acknowledge that ABCTracker is, like all existing systems, not perfect, and that several aspects of the system could be improved or further investigated in future works. Disadvantages of the current system include:

\begin{itemize}[label=\faChevronCircleDown]
    \item During the marking phase, some object exhibit shapes and poses that are less suitable for the three-click user marks. The three-click procedure works well for a variety of objects (such as ants, and termite) and performs adequately for some objects (e.g., zebrafish), but lacks the flexibility to handle objects with intricate poses (e.g., snakes). This shortcoming can be addressed in future releases by adding an optional “painting” mode during the marking phase. 
    
    \item Currently, the marking phase is required for every video. Some other systems may be able to reuse parameters determined in previous videos. 
    
    \item The default tracking algorithm is prone to making far more FN associations (i.e., fragmented trajectories) when objects exhibit very rapid changes in velocity. Additionally, the default tracking algorithm does not benefit from longer videos as much as some algorithms do in the related works. For example, idTracker.ai is able to construct more discriminative fingerprints when it has access to more training examples \cite{idtrackerai} (i.e., longer videos). 
    
    \item The tracking phase takes a decent amount of time relative to several existing systems. Automatic chunking does provide speedups on longer videos through parallel processing of chunks, but applying tracking on a chunk still requires several hours. Note that the client’s machine is not burdened during the tracking phase because computation is performed on the server. This means that while the server performs tracking on a video, the user is free to create, mark, initiate tracking, or even perform corrections on other videos.
    
    \item A potential downside of guided correction mode concerns how much information is hidden from the user. Currently, as soon as a review has been answered the next one is presented. Some users have reported that they would like to know more about how their review answer affected the tracks. As future work, we would like to investigate ways of including additional information during guided reviews without making guided mode more complicated or sacrificing efficiency.  
    
\end{itemize}

Other future works include: reducing the number of user-marked frames during the marking phase, and improving inference during guided correction mode (e.g., better selection of key-frames). Finally, we want to leverage the modular structure of the tracking framework further by including other tracking algorithms in addition to the default tracker described in this work.

\bibliographystyle{natbib}

    
\end{document}

%% file: SystemOverview.tex
    \begin{figure*}[ht!]
        \centering
            \includegraphics[width=\textwidth]{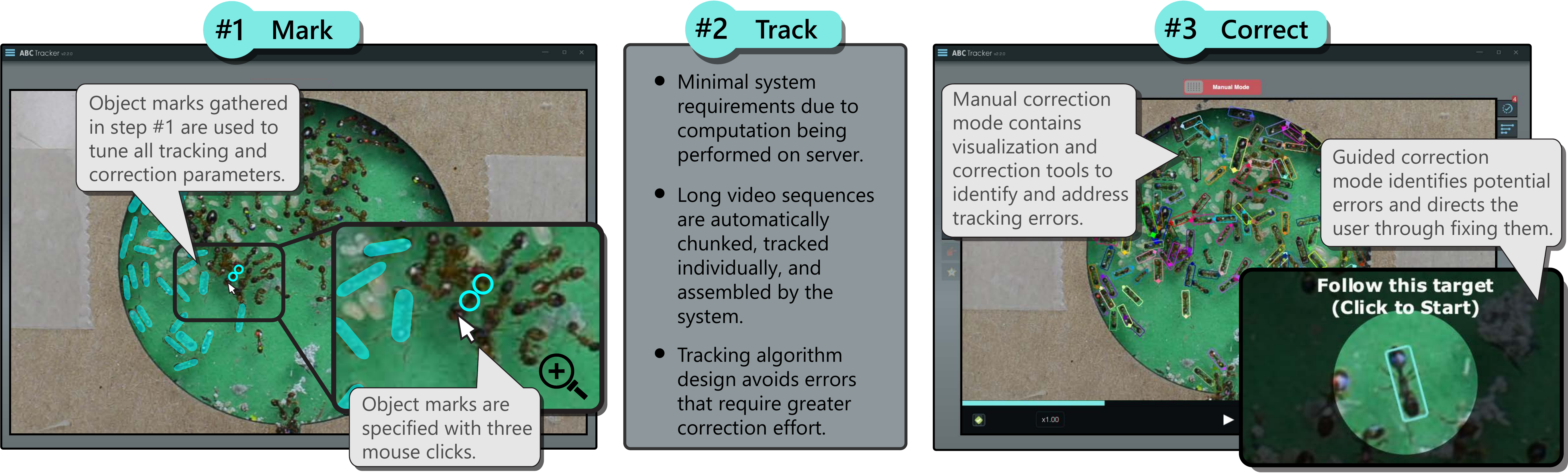}
                \caption{ABCTracker is a multi-object tracking system that operates in three phases -- Marking, Tracking, and Correction. In the first phase, the user marks objects in a few of the video frames using a simple three-click procedure (head, middle, tail/foot - or vise versa). The marks created in this initial phase provide everything needed to tune all parameters used during the tracking and correction phases. After the marking phase, automated tracking begins its execution on the remote server, thus allowing individuals to mark, track, and correct other recordings in the meantime. Long video sequences are automatically chunked, tracked individually, and assembled by the system. This enables the system to process the video chunks in parallel. In the last phase, the user is provided with several tools needed to address any errors in the tracking results.}
            \label{Teaser}
    \end{figure*}

%% file: Algorithm_setup_comparison.tex
    \begin{figure*}[ht!]
    \centering
        \includegraphics[width=\textwidth]{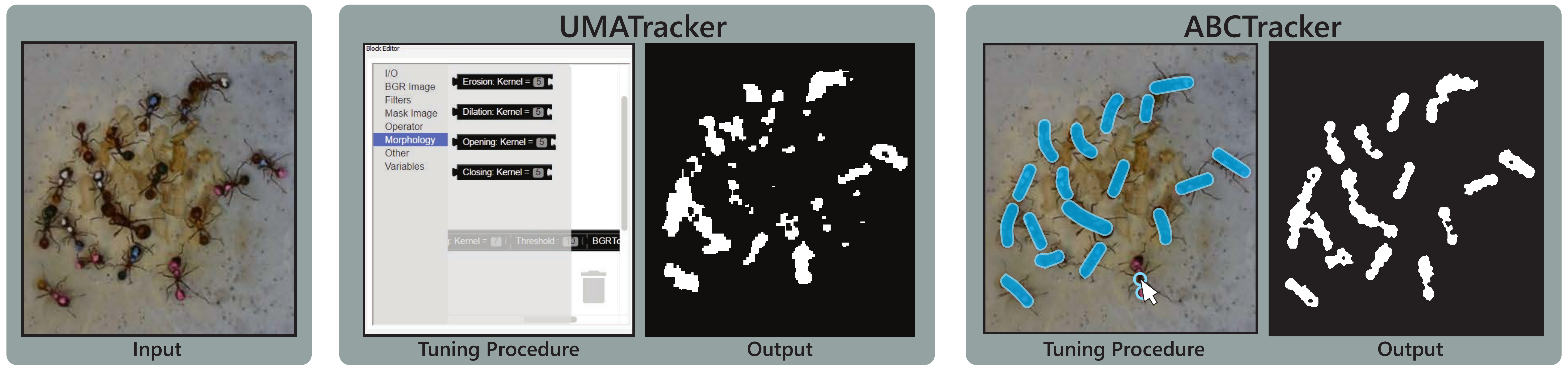}
            \caption{An illustration comparing typical parameter tuning procedures against the proposed approach. Existing tracking systems request that parameter values be defined directly (e.g. \cite{idtrackerai, idtracker, toxtrac, biotracker, ctrax}), and in some cases require the user to define portions of the tracking pipeline as well \cite{SwisTrack, umatracker}. An advantage of these approaches is that parameters and their values can potentially be applied to multiple recordings. The drawbacks concerns technical knowledge requirements (e.g. assumes knowledge of relationships between foreground properties and the algorithms ability to track) and encourages a time-consuming procedure of trial and error. The proposed system uses an intuitive object annotation approach to tune all tracking and correction related parameters. This allows the system to optimize the image processing pipeline automatically (particle swarm optimization, section-\ref{foregroundsection}) in addition to providing other useful facts about the recorded targets (e.g. size distribution). In this example, roughly twice as much time was spent tuning parameters in UMATracker \cite{umatracker} (middle, uses a graphical programming approach to define both an image processing pipeline along with the pipeline's parameter values) than was spent marking objects in ABCTracker. Note that ABCTracker was able to produce more complete foreground blobs with less noise.}
        \label{UserMarks}
    \end{figure*}

%% file: AlgorithmFlowChart.tex
    \begin{figure}[t!]
        \centering
            \includegraphics[width=\columnwidth]{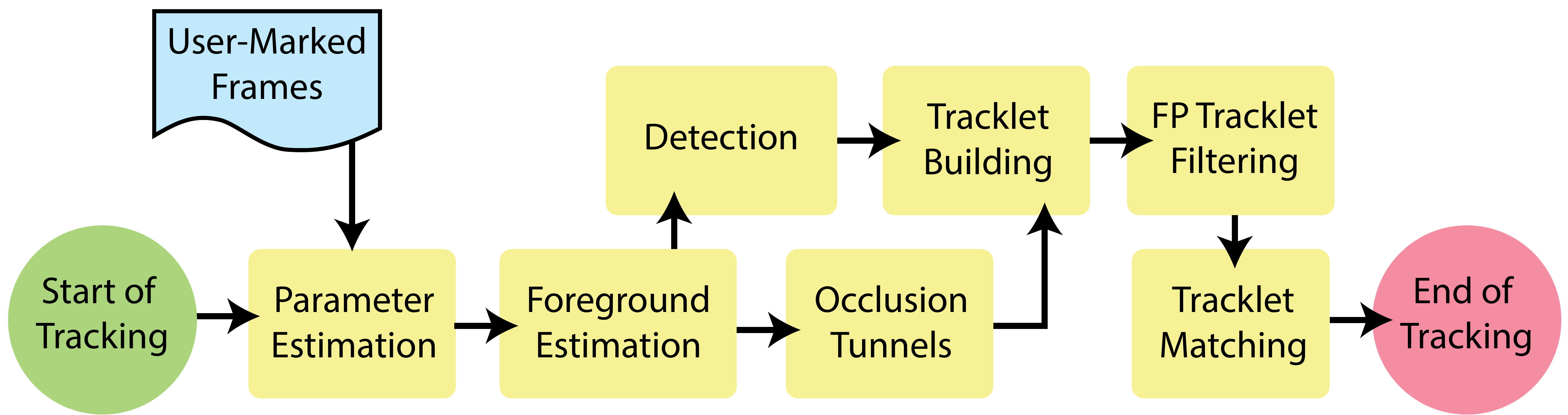}
                \caption{An overview diagram of the default algorithm used during the tracking phase. Expected inputs to the algorithm are the user-marked frames gathered during the marking phase. First, tracking parameters are determined based on the user-marks. Next, foreground classification is performed (section-\ref{foregroundsection}) and the results are used both as region proposals during detection (section-\ref{detectionsection}) and to construct occlusion tunnels \cite{fascianoTunnels} (section-\ref{trackletbuildingsection}). Tracklet building accepts the detection responses and occlusion tunnels, using them to build an initial set of high-confidence trajectory fragments (i.e., tracklets) (section-\ref{trackletbuildingsection}). Once the initial tracklet set is determined, a detection confidence classifier is trained and applied. The system then attempts to filter false positive tracks based on detection confidence. Finally, tracklet matching is iteratively performed using prediction agreements between independent sequential trackers (section-\ref{matchingsection}). For videos of sufficient length, chunking is performed automatically such that the algorithm is applied to each chunk individually and stitched together afterwards (section-\ref{chunkingsection}).}
            \label{Flowchart}
    \end{figure}

%% file: Forward-backward_GAPF.tex
    \begin{figure*}[ht!]
    \centering
        \includegraphics[width=\textwidth]{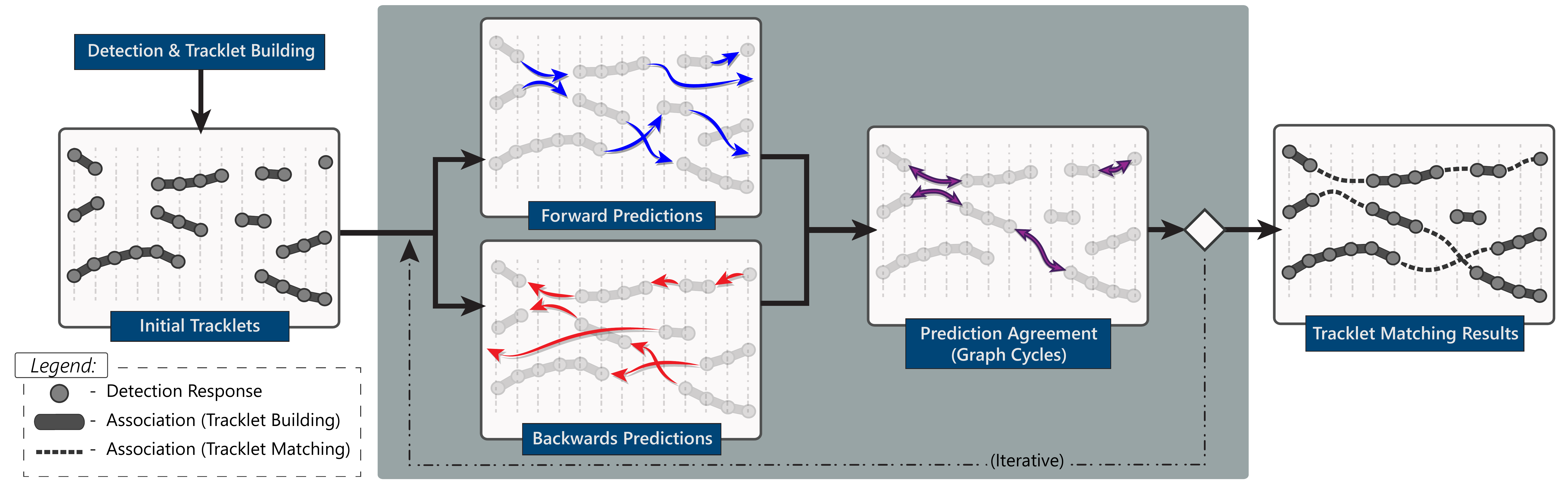}
            \caption{Overview of the tracklet matching algorithm (section-\ref{matchingsection}) given the initial set of tracklets from tracklet building (section-\ref{trackletbuildingsection}). Sequential trackers, initiated from tracklet end points, are run both forward (blue arrows) and backward (red arrows). A cycle in the association graph -- representing an agreement between forward and backward tracker predictions in a single iteration (illustrated as purple arrows) -- are iteratively determined to conservatively match tracklets.}
        \label{IterativeMatching}
    \end{figure*}

%% file: Large_GuidedInferenceExample.tex
    \begin{figure*}[ht!]
        \centering
            
            \includegraphics[width=\textwidth]{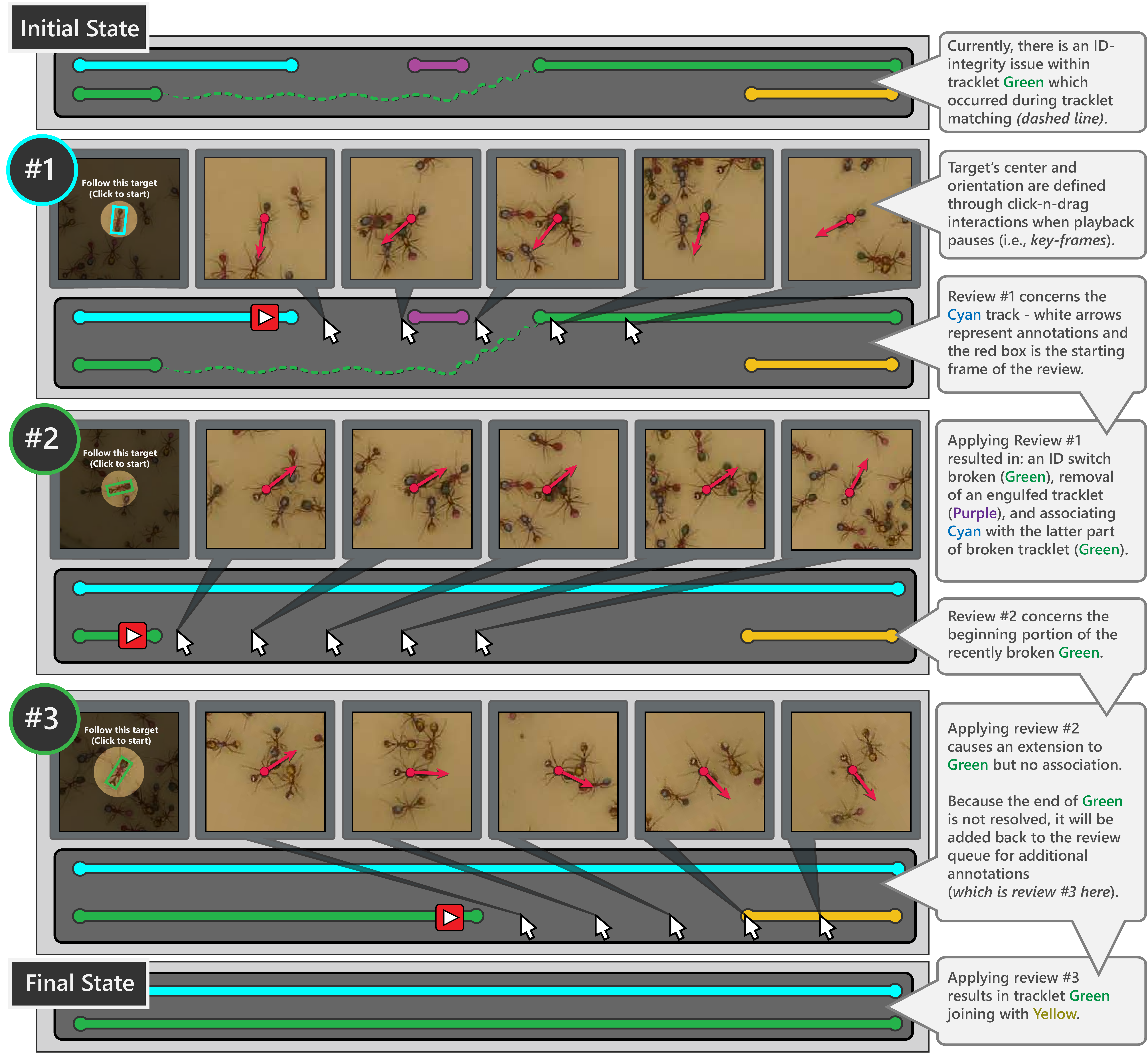}
                \newline
                \caption{An illustration of guided mode reviews and the resulting state of the tracklets once inference is performed on review answers. Here, five rows are shown: the first and last rows depict the initial and final state of the tracklets, the three middle rows (numbered \#1--\#3) portray reviews and their answers.  Each colored line within the tracking-state drawings represents a tracklet (e.g., horizontal lines colored cyan, green, yellow, and purple). Two tracklets (i.e., lines) that are vertically aligned with one another represent a correct associations. The sequence of images within each of the review blocks numbered \#1--\#3 shows the key-frames presented to the user (i.e. playback pause points) and the how the user specifies the target's position and orientation (red arrows). The initial state of the tracklets contains several FN associations (i.e. gaps between vertically aligned lines) and one ID-integrity issue (i.e., FP association made during tracklet matching -- represented by the dashed green line). The ID-integrity issue is corrected while addressing the premature termination on the cyan track (Review \#1). Review \#1 also shows an example of removing an engulfed track (purple track/line). Reviews \#2 and \#3 handle the remaining FN association and demonstrate how additional annotations are sometimes required to establish an association.    
                }
            \label{GuidedInference_Large}
    \end{figure*}

%% file: Tracking_Gallery.tex
    \begin{figure*}[ht!]
        \centering
            \includegraphics[width=\textwidth]{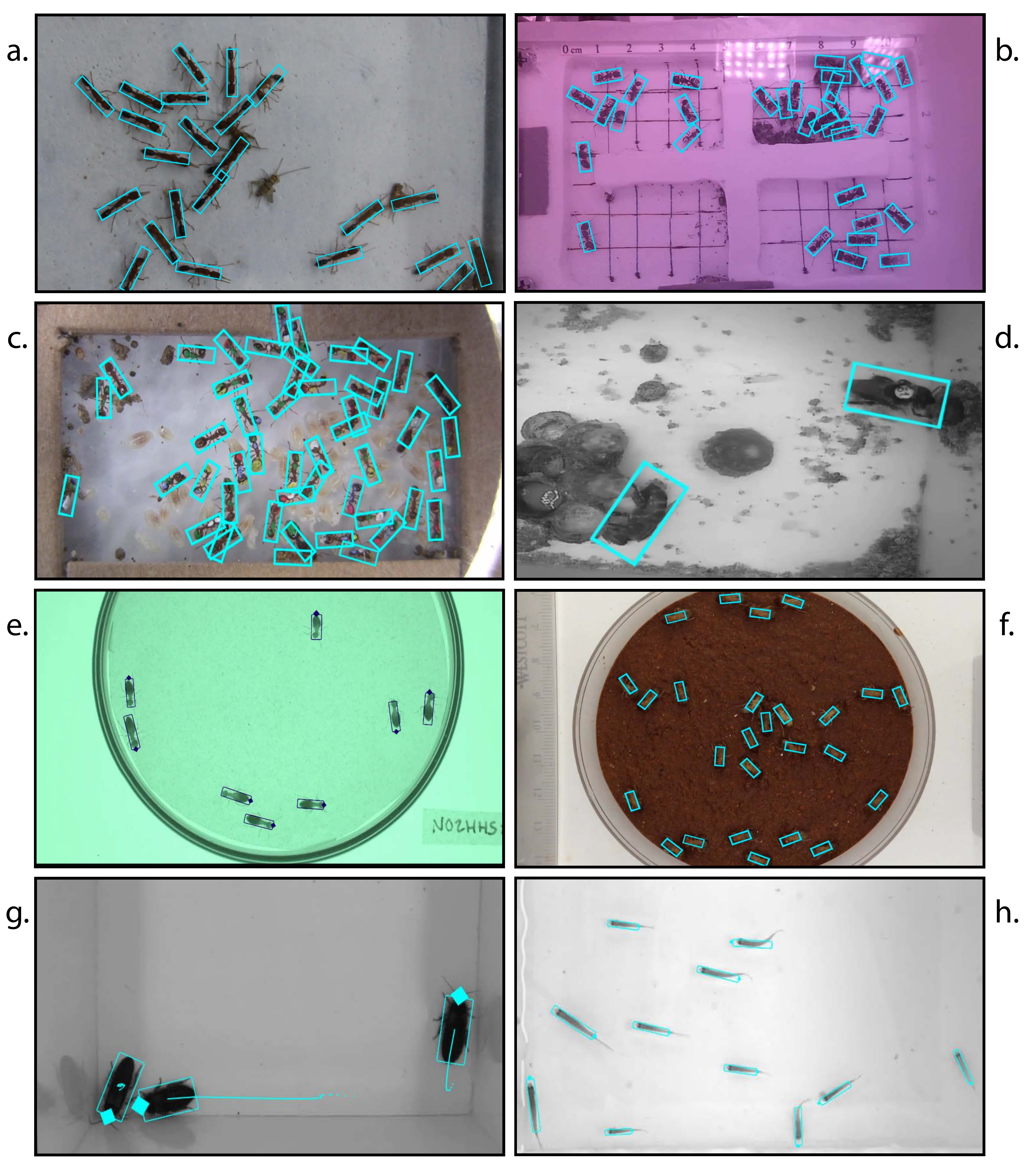}
                \caption{Examples of tracking results gathered using ABCTracker. (a-c) Ants under various lighting and background conditions, (d) bees which exhibit large pose variations in a scene with difficult background textures, (e) termites recorded with back-lighting (f) termites that shift dirt slowly over time as they move, (g) cockroaches, and (h) zebrafish. }
            \label{Gallery}
    \end{figure*}

%% file: PossibleErrorsTable.tex
    \begin{figure*}[ht]
        
        \caption{List of tracking failures, their descriptions, Operation that corrects them in manual mode, and whether errors of that type are handled in guided correction mode.}
        \label{Error-type-list}

            \begin{center}
                \begin{tabular}{ || p{1.3in} | p{2.8in} | p{1.0in} | p{0.8in}||}
                    \hline
                    
                    Tracking Error &
                        Description of Tracking Error &
                        Addressed within Guided Mode &
                        Addressed by Manual Op. \\ \hline \hline
                    
                    FN Association &
                        Tracking fails to associate multiple tracks covering a single object. &
                        Yes &
                        \raisebox{-0.6\totalheight}{\includegraphics[scale=1.5]{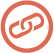}}
                            ~\raisebox{-0.5\totalheight}{Join} 
                        \\ \hline
                    
                    ID Switches & 
                        An incorrect association that causes a track to switch from one target 
                            to another. &
                        Yes &
                        \raisebox{-0.6\totalheight}{\includegraphics[scale=1.5]{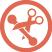}}
                            ~\raisebox{-0.5\totalheight}{Break} 
                        \\ \hline
                    
                    False Tracks (FP) &
                        A track never covers an object in the video, or does but less appropriately 
                            than other track. &
                        Yes &
                        \raisebox{-0.6\totalheight}{\includegraphics[scale=1.5]{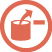}}
                                ~\raisebox{-0.5\totalheight}{Remove} 
                        \\ \hline
                    
                    Lost track &
                        A track covering an object ends prematurely and there is no correct association that would extent it. &
                        Yes, if extension is forward in time  &
                        \raisebox{-0.6\totalheight}{\includegraphics[scale=1.5]{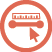}}
                                ~\raisebox{-0.5\totalheight}{Adjust} 
                        \\ \hline
                    
                    Localization errors &
                        A track does not accurately represent the state of the target it covers (location, 
                            orientation, or size) &
                        Some cases &
                        \raisebox{-0.6\totalheight}{\includegraphics[scale=1.5]{Images/Adjust-Icon.png}}
                                ~\raisebox{-0.5\totalheight}{Adjust} 
                        \\ \hline
                    
                    Missed Objects (FN) &
                        An object is never covered by a track at any point in the video. &
                        No &
                        \raisebox{-0.6\totalheight}{\includegraphics[scale=1.5]{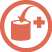}}
                                ~\raisebox{-0.5\totalheight}{Add} 
                        \\ \hline
                    
                    \hline
                \end{tabular}
        \end{center}
    \end{figure*}

%% file: Corrections_Interface.tex
    \begin{figure*}[ht!]
        \centering
            \includegraphics[width=\textwidth]{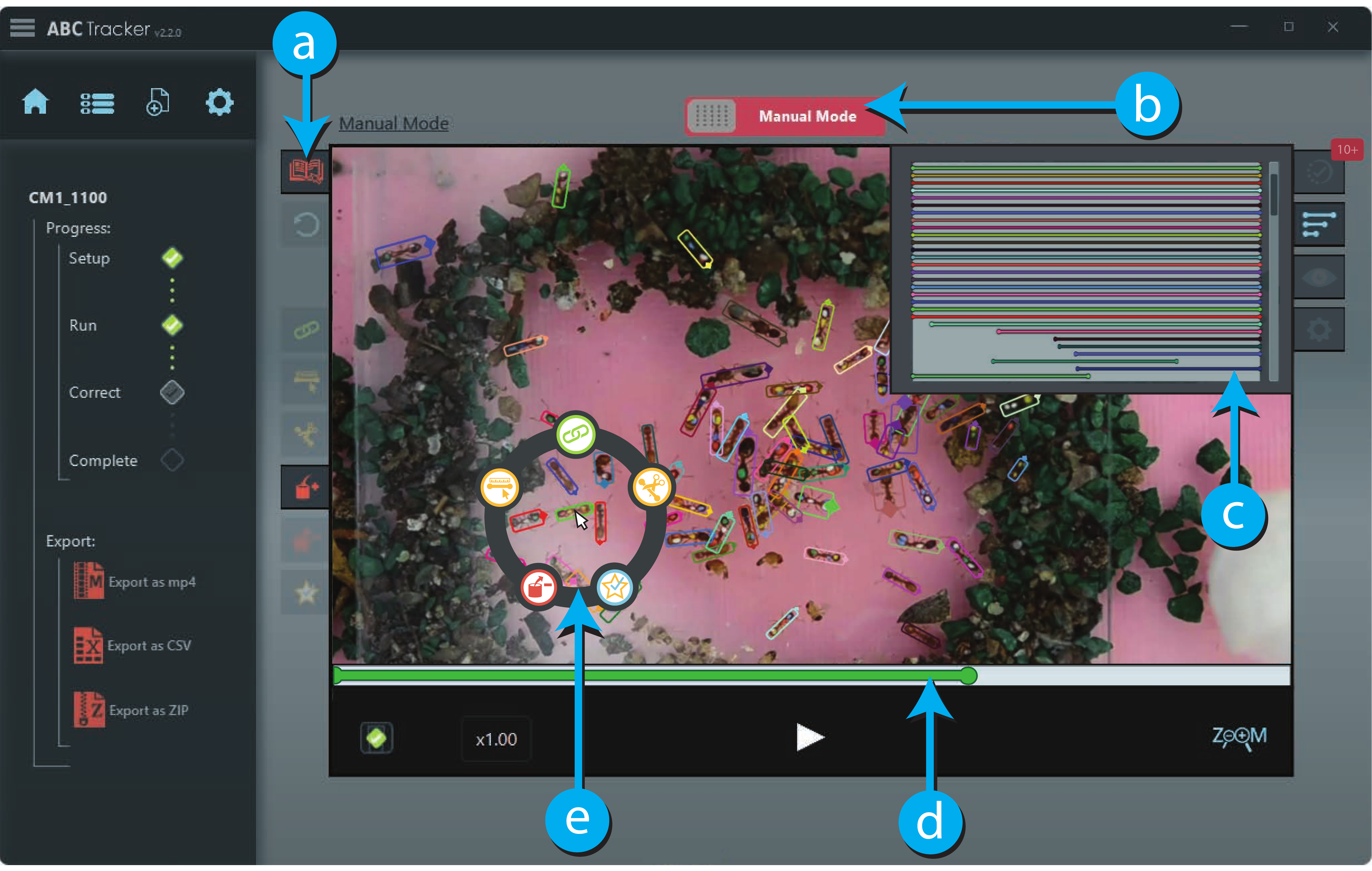}
                \caption{
                An overview of the correction features for a tracking process in manual correction mode. The right click menu (e) shows the available operations for the track selected. Several features are included to aid with performing corrections, for example: (d) above the video playback bar sits a linear representation of the selected track (green colored -- matches the color of the selected track) which depicts the temporal extent of the track and (c) an interactive tab showing the temporal extent of all tracks in the video. Switching between guided and manual correction modes can be done at any time with the toggle button (b). When users encounter a feature for the first time, tutorial dialogs are shown automatically to explain how and when to use that feature. All tutorials can be viewed again by clicking (a).}
            \label{ManualVisualizationFeatures}
    \end{figure*}

%% file: GuidedSimulations_Baseline281.tex
    \begin{figure*}
    	\begin{minipage}{\textwidth}
    		\centering
    		\noindent\makebox[\linewidth]{\rule{\textwidth}{0.4pt}}
    		
    		\includegraphics[width=1.02\linewidth]{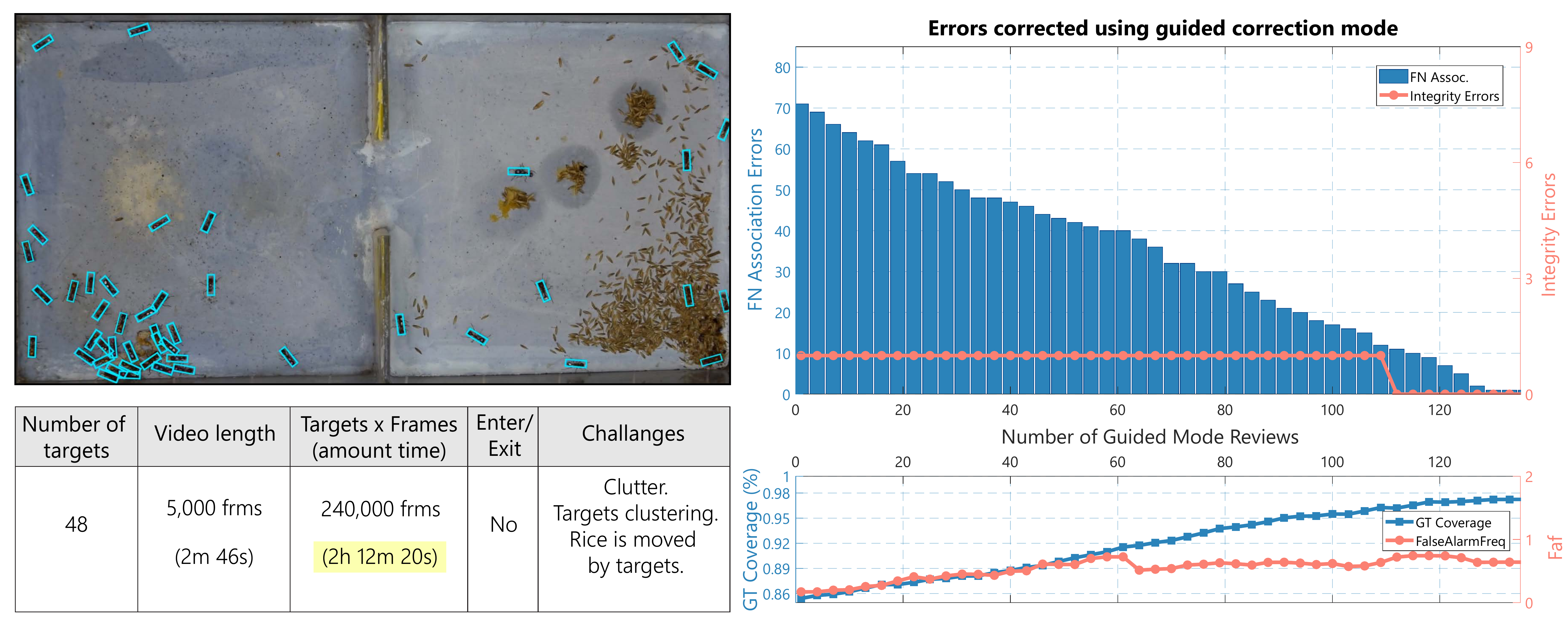}

    		\centering
    		
    		\begin{tabular}{|c|c|c|c|c|c|c|c|c|c|}
                \hline
                ~~~~~~~~~~~~~~~~~~~~~~~~~~~~~~~~~~~~~~~~ &
                GT Cov. &
                FAF &
                Avg Pos. Error &
                FN Assoc. &
                ID Integ. &
                ID Sw. &
                MT &
                PT &
                ML \\
                \hline
                \hline
                
                Tracklet Building Output &
                0.55 & 0.08 & 1.79 &
                2219 & 0 & 0 &
                6 & 36 & 6
                \\ \hline
                
                Tracklet Matching Output &
                0.85 & 0.17 & 2.57 &
                76 & 1 & 1 &
                37 & 9 & 2
                \\ \hline
                
                Guided Corrections Output &
                0.97 & 0.641 & 2.51 &
                0 & 0 & 0 &
                47 & 1 & 0
                \\ \hline
    
            \end{tabular}
            
    	\end{minipage}
    	
    	\noindent\makebox[\linewidth]{\rule{\textwidth}{0.4pt}}
    	\caption{Performance on the \textit{"OpenDivide\_1"} dataset. The \textit{Targets x Frames} column relates how long it would take to follow each target individually throughout the video, and in several ways, represents the quantity of tracking data to be gathered.}
    	\label{GuidedModeEval_Baseline281}

    \end{figure*}

%% file: GuidedSimulations_CM1_0693.tex
    \begin{figure*}
    
    	\begin{minipage}{\linewidth}
    		\centering
    		
    		\noindent\makebox[\linewidth]{\rule{\textwidth}{0.4pt}}
    		
    		\includegraphics[width=1.02\linewidth]{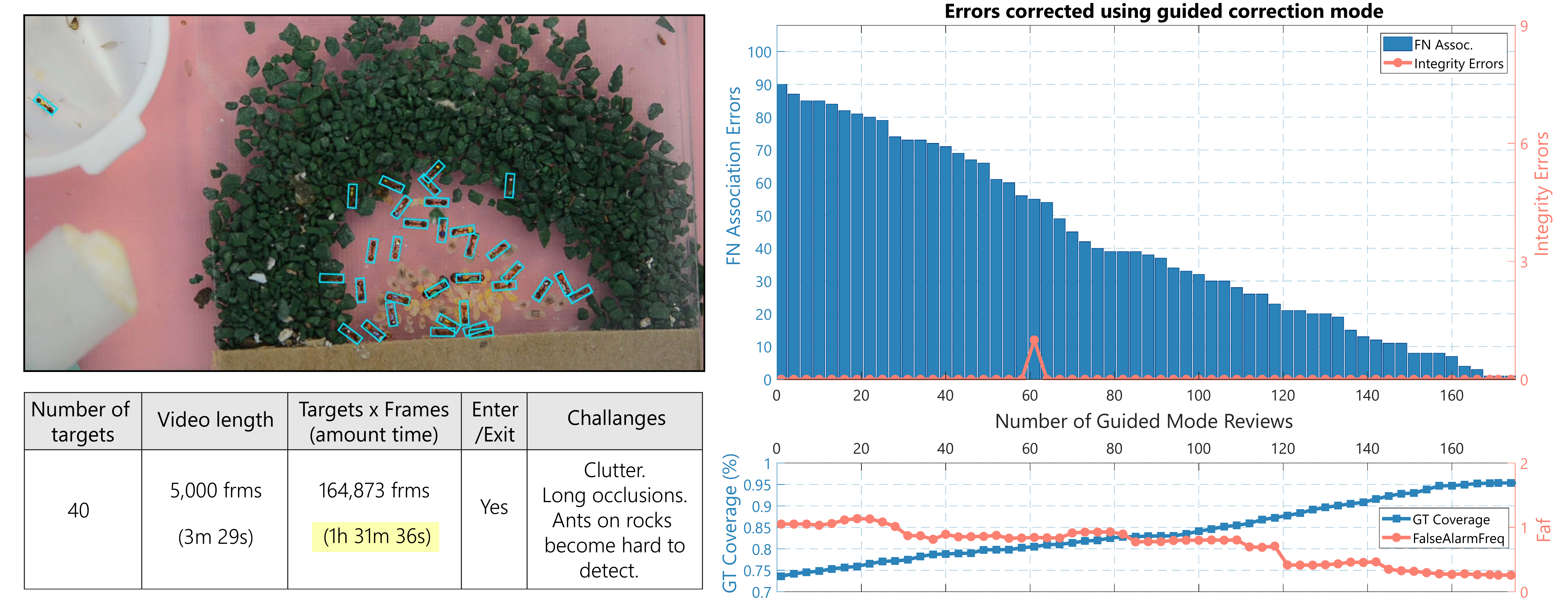}

    		\centering
    		
    		\begin{tabular}{|c|c|c|c|c|c|c|c|c|c|}
                \hline
                ~~~~~~~~~~~~~~~~~~~~~~~~~~~~~~~~~~~~~~~~ &
                GT Cov. &
                FAF &
                Avg Pos. Error &
                FN Assoc. &
                ID Integ. &
                ID Sw. &
                MT &
                PT &
                ML \\
                \hline
                \hline
                
                Tracklet Building Output &
                0.628 & 0.32 & 2.99 &
                1355 & 0 & 0 &
                19 & 15 & 6
                \\ \hline
                
                Tracklet Matching Output &
                0.736 & 1.11 & 3.17 &
                114 & 0 & 0 &
                23 & 11 & 6
                \\ \hline
                
                Guided Corrections Output &
                0.946 & 0.51 & 6.36 &
                0 & 0 & 0 &
                37 & 1 & 2
                \\ \hline
                
            \end{tabular}

    	\end{minipage}

    	~\\~\\

    	\noindent\makebox[\linewidth]{\rule{\textwidth}{0.4pt}}
    	\caption{Performance on the \textit{"PinkArena\_1"} dataset. The \textit{Targets x Frames} column relates how long it would take to follow each target individually throughout the video, and in several ways, represents the quantity of tracking data to be gathered.} 

    	\label{GuidedModeEval_CM1_0693}
    \end{figure*}

%% file: ManualAnnotTable_Baseline281.tex
    \begin{figure*}
        \centering

            \caption{\textbf{(a)} Manual annotation, simulated using ground truth Annotations are specified every \textit{StepSz} frames with linear interpolation. (Note: this is not referring to ABCTracker's Manual correction mode.)}
            \begin{tabular}{||c|c|c|c|c|c|c|c|c|c|c||} 
                \hline
                ~~StepSz~~~ & \#-Annotations & Playback Time & Est. Time & GT Cov. & MT & PT & ML & IdIntg. & IDS & FAF  
                \\ \hline

                	 20  & 12,048 &  2h 12m & 7h 16m & 0.99 
                		& 48 & 0 & 0 
                		& 0 & 0 & ~~0.16
                		\\ \hline

                \rowcolor{LightCyan}
                    30 & 8,064 &  2h 12m & 5h 36m & 0.97
                    	& 48 & 0 & 0 
                    	& 0 & 2 & ~~1.22
                    	\\ \hline
                    	
                	40 & 6,048 &  2h 12m & 4h 46m & 0.92
                		& 47 & 1 & 0 
                		& 0 & 2 & ~~3.41
                		\\ \hline

                	 60 & 4,080 &  2h 12m & 3h 56m & 0.82
                		& 25 & 23 & 0 
                		& 0 & 12 & ~~8.61
                		\\ \hline

                	 80 & 3,072 &  2h 12m & 3h 31m & 0.73 
                		& 16 & 32 & 0 
                		& 3 & 45 & 12.95
                		\\ \hline

                	 100 & 2,448 &  2h 12m & 3h 16m & 0.66
                		& 12 & 36 & 0 
                		& 6 & 86 & 16.11
                		\\ \hline

                	 300 & 864 &  2h 12m & 2h 36m & 0.39 
                		& 7 & 23 & 18 
                		& 103 & 313 & 28.92
                		\\ \hline

                	 500 & 528 &  2h 12m & 2h 28m & 0.30 
                		& 4 & 20 & 24 
                		& 225 & 395 & 33.41
                		\\ \hline

                	 1000 & 288 &  2h 12m & 2h 22m & 0.21 
                		& 2 & 14 & 32 
                		& 202 & 350 & 37.44
                		\\ \hline
                
            \end{tabular}

        ~\\
            \caption{\textbf{(b)} Guided mode corrections, simulated using ground truth. Performance reported after \textit{\#-Reviews} have been answered.}
            ~\\
            \begin{tabular}{||c|c|c|c|c|c|c|c|c|c|c||}
                \hline
                \#-Reviews & \#-Annotations & Playback Time & Est.Time & GT Cov. & MT & PT & ML & IdIntg. & IDS & FAF  
                \\ \hline
                
                	30 & 300 & 6m & 14m & 0.88 
                		& 39 & 9 & 0 
                		& 1 & 1 & ~~0.45
                		\\ \hline
                		
                	60 & 600 & 12m & 29m & 0.92 
                		& 41 & 7 & 0 
                		& 1 & 1 & ~~0.72
                		\\ \hline
                		
                	90 & 900 & 18m & 43m & 0.95 
                		& 45 & 3 & 0 
                		& 1 & 1 & ~~0.64
                		\\ \hline
                		
                	120 & 1,197 & 23.9m & 58m & 0.97
                		& 47 & 1 & 0 
                		& 0 & 0 & ~~0.74
                		\\ \hline
                		
            	\rowcolor{LightCyan}
                	131 (All) & 1,247 & 24.7m & 1h ~1m & 0.97 
                		& 47 & 1 & 0 
                		& 0 & 0 & ~~0.64
                		\\ \hline

            \end{tabular}

        ~\\~\\~\\
        \caption{Comparison between manual annotation and ABCTracker's guided correction mode on the dataset \textit{"OpenDivide\_1"} (see figure-\ref{GuidedModeEval_Baseline281} for details). Table (a) represents the traditional approach to gathering trajectory data by manually annotating each target's position every set interval of frames (\textit{StepSz}). Table (b) shows the results of simulated guided corrections using ground truth (outlined in section-\ref{EvaluationSection}). Within both tables, the amount of time spent watching the video is reported in the \textit{Playback Time} column, and the \textit{\#-Annotations} column represents the number of times a click-n-drag annotation is supplied while playback is paused.  An estimated time that includes video playback, the number of annotations (constant time), and changing between review/targets (constant time) is reported in the \textit{Estimated Time} column. Please note that the values within the \textit{Estimated Time} columns may be underestimates. Still, since both methods use common parameters for estimating time (see \ref{EvalCorrection}), they should be representative of the relative difference between the two approaches. The traditional approach of manual annotation  achieves similar ground truth coverage as answering all guided corrections (final row of (b)) when the key-frame step size equals 30 (cyan row in (a)) but requires 1.78 hours of additional playback time and 6,817 additional annotations.
        }
        \label{manualannotationcomparison}
    \end{figure*}